\documentclass[10pt,twocolumn,letterpaper]{article}

\usepackage{cvpr}      

\usepackage{booktabs}  
\usepackage{makecell}  

\newcommand{\figref}[1]{Figure~\ref{#1}}
\newcommand{\tabref}[1]{Table~\ref{#1}}
\newcommand{\secref}[1]{Section~\ref{#1}}

\usepackage{times}
\usepackage{helvet}
\usepackage{courier}
\usepackage[hyphens]{url}
\usepackage{graphicx}
\usepackage{amsmath}
\usepackage{amssymb}
\usepackage{booktabs} 
\usepackage{caption}
\usepackage{subcaption} 
\usepackage{algorithm}
\usepackage{algorithmic}

\usepackage{relsize}

\definecolor{cvprblue}{rgb}{0.21,0.49,0.74}
\usepackage[pagebackref,breaklinks,colorlinks,allcolors=cvprblue]{hyperref}

\title{Stand-In: A Lightweight and Plug-and-Play Identity Control \\ for Video Generation}

\author{Bowen Xue$^{1,*}$\quad  Zheng-Peng Duan$^{2,1,*}$ \quad Qixin Yan$^{1,\#}$\quad Wenjing Wang$^1$ \\ Hao Liu$^1$\quad Chun-Le Guo$^2$\quad Chongyi Li$^2$\quad Chen Li$^1$\quad Jing LYU$^1$ \\
$^1$WeChat Vision, Tencent Inc. \quad
$^2$VCIP, CS, Nankai University\\
{\tt\small \{bowenxue2005,adamduan0211\}@gmail.com,}\\
{\tt\small \{qixinyan,augustawang,leweshaoliu,chaselli,eckolv\}@tencent.com,} \\
{\tt\small \{guochunle,lichongyi\}@nankai.edu.cn,} \\
{\small {\url{https://github.com/WeChatCV/Stand-In}}}}

\begin{document}

\twocolumn[{%
\renewcommand\twocolumn[1][]{#1}%
\maketitle
\vspace{-10mm}
\begin{center}
    \centering
    \captionsetup{type=figure}
    \includegraphics[width=\linewidth]{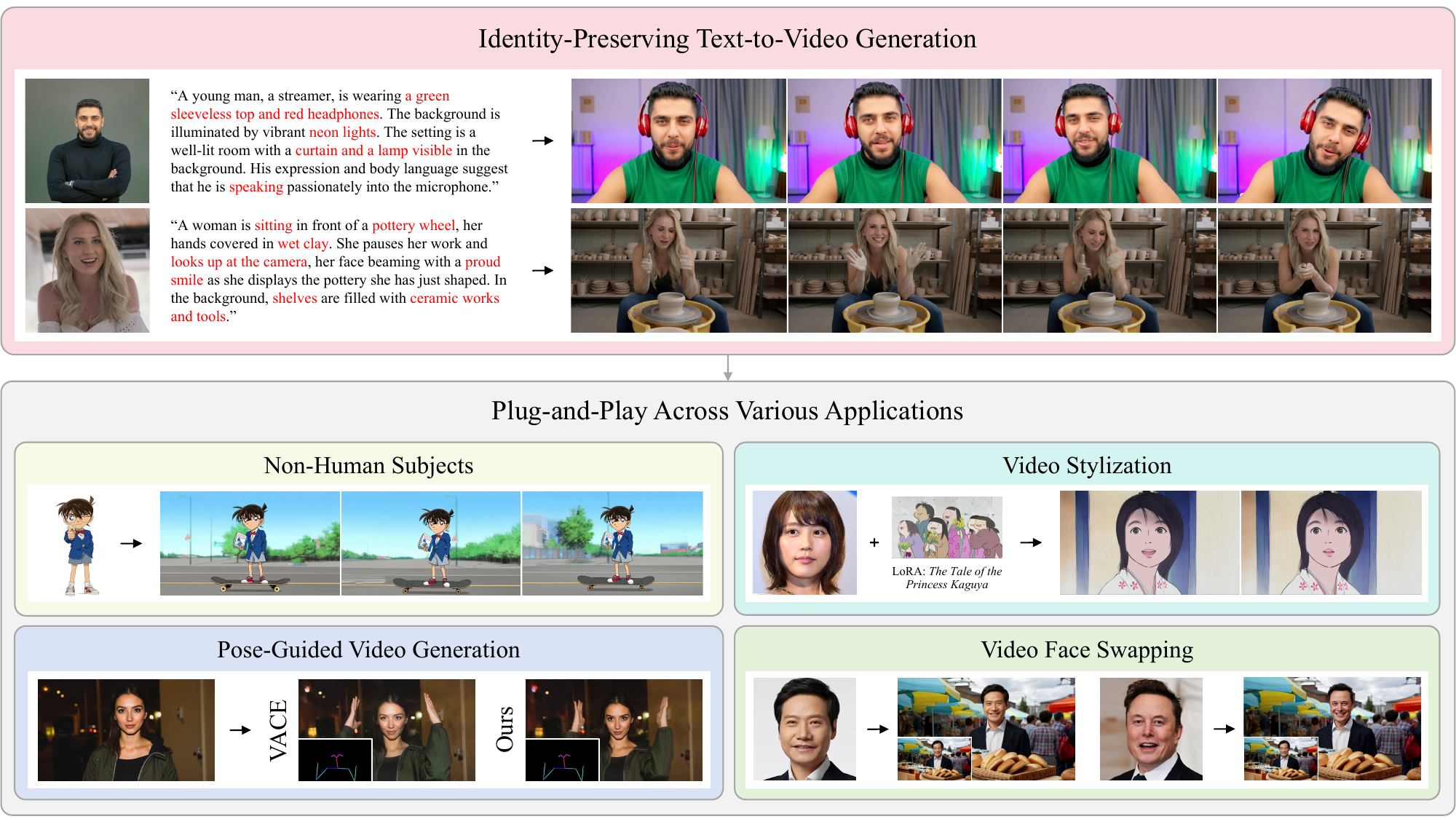}
    \captionof{figure}{Given a reference image, our method generates videos with strong identity preservation. Furthermore, the framework's plug-and-play design enables seamless integration into diverse applications for enhanced identity consistency.}\label{fig:teaser}
    \vspace{4mm}
\end{center}%
}]
\let\thefootnote\relax\footnotetext{$^{*}$ Equal Contribution.}
\let\thefootnote\relax\footnotetext{$^{\#}$ Corresponding Author.}

\begin{abstract}
Generating high-fidelity human videos that match user-specified identities is important yet challenging in the field of generative AI.
Existing methods often rely on an excessive number of training parameters and lack compatibility with other AIGC tools.
In this paper, we propose \textbf{Stand-In}, a lightweight and plug-and-play framework for identity preservation in video generation.
Specifically, we introduce a conditional image branch into the pre-trained video generation model.
%
%
%
Identity control is achieved through restricted self-attentions with conditional position mapping.
Thanks to these designs, which greatly preserve the pretrained prior of the video generation model, our approach is able to outperform other full-parameter training methods in video quality and identity preservation, even with just $\sim$1\% additional parameters and only 2000 training pairs.
Moreover, our framework can be seamlessly integrated for other tasks, such as subject-driven video generation, pose-referenced video generation, stylization, and face swapping.
%
\end{abstract}    
\vspace{-5mm}
\section{Introduction}
\label{sec:intro}
With the rapid advancement of diffusion models~\cite{NEURIPS2020_4c5bcfec,podell2023sdxl,Peebles_2023_ICCV}, video generation~\cite{opensora,opensora2,kong2024hunyuanvideo,hong2022cogvideo} has become a 
pivotal aspect of generative AI.
Among its diverse applications, identity-preserving video generation holds profound significance.
The goal of this task is to generate high-quality videos that consistently maintain the identity of a given reference image containing a face. 
It has widespread utility across film, advertising and gaming industries, \etc

Existing methods can be roughly classified into two categories. Early methods~\cite{ID-Animator,consistid} use an explicit face encoder for identity feature extraction, while recent methods~\cite{hu2025hunyuancustom,liu2025phantom} fully train the diffusion transformer.
However, face-encoder-based methods lack flexibility and struggle to capture fine facial details essential for high-quality video generation.
The full-parameter training methods require huge training resources and lack compatibility with other applications. 
Achieving robust identity preservation in a lightweight way remains critical yet challenging.

\begin{figure}[t]
\centering
\includegraphics[width=\columnwidth]{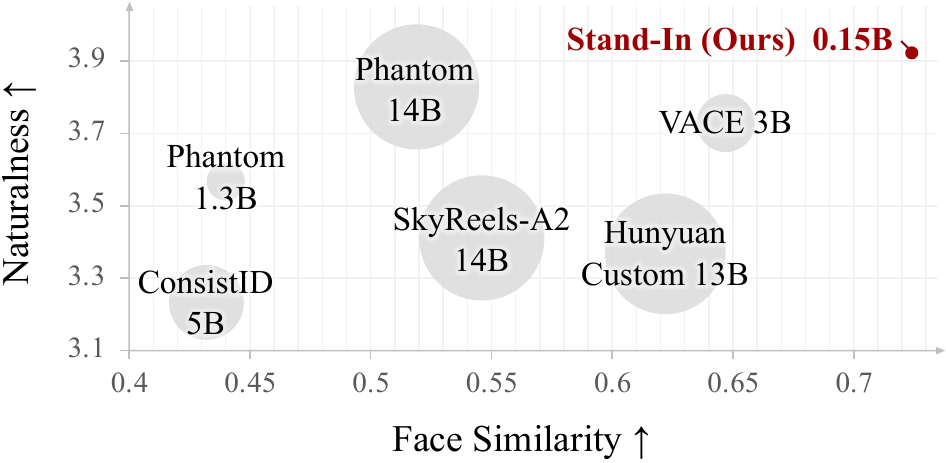}
\caption{Comparison with SOTA identity-preserving video generation methods. The size of bubbles represents the number of need-to-train parameters for identity preservation. Our approach achieves the highest performance in both face similarity and naturalness, while utilizing the fewest parameters.}
\label{fig:bubble}
\vspace{-4mm}
\end{figure}

To overcome these limitations, we leverage the pre-trained VAE from the video generation model itself, enabling the conditional image to be mapped directly into the same latent space as the video.
This approach naturally utilizes the model's inherent capabilities to extract rich and detailed facial features, offering a more integrated and effective solution.
%
%
Specifically, we employ restricted self-attention with conditional position mapping to merge the features of the reference image into the video.
On the one hand, by preserving the core functionality of self-attention and the pretrained prior of the video generation model, our method achieves the highest facial similarity and naturalness in identity-preserving video generation with only $\sim$1\% additional parameters and 2000 training pairs, which is shown in Figure~\ref{fig:bubble}.
On the other hand, our method does not alter the architecture of the main video generation model and thus can be used in a plug-and-play manner for other applications.
%
As shown in Figure~\ref{fig:teaser}, our framework can be extended to various tasks, including subject-driven generation, video stylization, and face swapping, all guaranteeing identity consistency.
Additionally, benefit from integrating compatibility with VACE~\cite{vace}, our approach significantly enhances facial similarity in pose-guided video generation.
Our main contributions can be summarized as follows:

\begin{itemize}[leftmargin=*]

\item We present Stand-In, a lightweight and plug-and-play framework designed for identity-preserving video generation. By incorporating and training just $\sim$1\% additional parameters, our approach achieves SOTA results in identity preservation, video quality, and prompt following.

\item To inject identity information without explicit face feature extractors, we introduce a conditional image branch to the video generation model. The image and video branches share information through restricted self-attention with conditional position mapping. With these designs, identity preservation can be learned well with a small dataset.

\item The proposed framework exhibits high compatibility and generalizability.
Although trained only on real-people data, our method generalizes to other subjects, such as cartoons and objects. Moreover, our method can be plug-and-play applied to other tasks, such as pose-guided video generation, video stylization and face swapping.

\end{itemize}

\vspace{-3mm}
\section{Related Work}
\label{sec:relatedwork}

\paragraph{Video Generation}

Current video generation models are predominantly built on diffusion frameworks~\cite{NEURIPS2020_4c5bcfec}, with an evolution in architecture from U-Net-based designs~\cite{blattmann2023stablevideodiffusionscaling} to DiT-based approaches~\cite{kong2024hunyuanvideo,wan2.1,ma2025latte,DiT,wang2023laviehighqualityvideogeneration,VideoCrafter2,T2V-Turbo}.
In the era of U-Net-based diffusion models, text-to-image (T2I) frameworks~\cite{rombach2021highresolution,podell2023sdxl} were extended to video generation by introducing 3D convolutions and temporal attention ~\cite{blattmann2023stablevideodiffusionscaling}. 
AnimateDiff~\cite{guo2023animatediff} further advanced this direction by reusing pre-trained text-to-image model weights to leverage their strong spatial generation capabilities by adding temporal layers.
Latte~\cite{ma2025latte} introduced a spatial-temporal separation mechanism, assigning distinct DiT blocks to process spatial and temporal information independently. This approach was later replaced by 3D full attention mechanisms, which offered more integrated processing.
CogVideoX~\cite{yang2024cogvideox} and HunyuanVideo~\cite{kong2024hunyuanvideo} combined 3D-VAE ~\cite{yu2023language} with MM-DiT~\cite{esser2024scaling} to enhance video generation capabilities. WAN2.1~\cite{wan2.1} employs a 3D-VAE and adopts a DiT backbone for denoising, injecting semantic information into the diffusion process through cross-attention.

\paragraph{Identity-Preserving Generation}

Prior to the advent of zero-shot identity-preserving algorithms, generating content with consistent identity typically relied on case-by-case fine-tuning.\cite{Dreambooth,Lora,textual_inversion, multi-concept, dreamvideo, customvideo, motionbooth}, In contrast, training-free identity-preserving image generation methods\cite{ye2023ip,shi2024instantbooth,guo2025pulid,wang2024instantid,he2024uniportrait,sara2025ipcompose,qian2024omni,han2024face,mou2025dreamounifiedframeworkimage}  offer a zero-shot personalization solution by integrating identity features into pre-trained foundation models. These methods typically introduce parameterized plug-in modules or adapters to adjust and inject identity features into the generation process. A popular solution is to use a Face Encoder to extract face embeddings and inject them into the generation process via cross-attention. For example, InstantID\cite{wang2024instantid} and PuLID\cite{guo2025pulid} can generate high-quality, identity-preserving images.

In the field of identity-preserving video generation~\cite{ma2025decoupled, zhang2025magictalk}, early methods commonly rely on explicit face encoders for facial feature extraction to generate videos with identity-preserving.  ID-animator \cite{ID-Animator} leverages a pre-trained text-to-video diffusion model in conjunction with a lightweight face adapter to encode ID-relevant embeddings from adaptable facial latent queries. ConsistID \cite{consistid}  aims to maintain identity consistency through frequency decomposition in diffusion transformer. Phantom \cite{liu2025phantom} can also preserve identity consistency in the human domain  as a unified subject-consistent video generation framework. HunyuanCustom \cite{hu2025hunyuancustom}
is a multi-modal customized video generation framework that emphasizes identity consistency while supporting diverse input modalities. By introducing advanced condition injection mechanisms and identity-preserving strategies, it achieves excellent performance in high-quality video generation. They employed a full fine-tuning for the diffusion transformer, requiring huge training resources.

\vspace{-2mm}
\section{Method}
\vspace{-1mm}
In this section, we first introduce the overall framework of the proposed method in~\secref{sec:overall}. Next, we detail the restricted self-attention mechanism in~\secref{sec:rsa} and conditional position mapping in~\secref{sec:cpm}. Finally, we present the data collection process in~\secref{sec:dataset}.
\vspace{-1mm}
\subsection{Overall Framework}
\vspace{-1mm}
\label{sec:overall}
To extract facial features, early methods~\cite{ID-Animator,consistid} rely on explicit face encoders, which lack flexibility and often fail to preserve fine facial details critical for high-quality reconstruction. In contrast, we propose using the pre-trained VAE from the video generation model.
This strategy maps the conditional image directly into the same latent space as the video and allows us to naturally take advantage of the built-in ability of the pre-trained video generation model to extract rich facial features.

\begin{figure}[t]
\centering
\includegraphics[width=\linewidth]{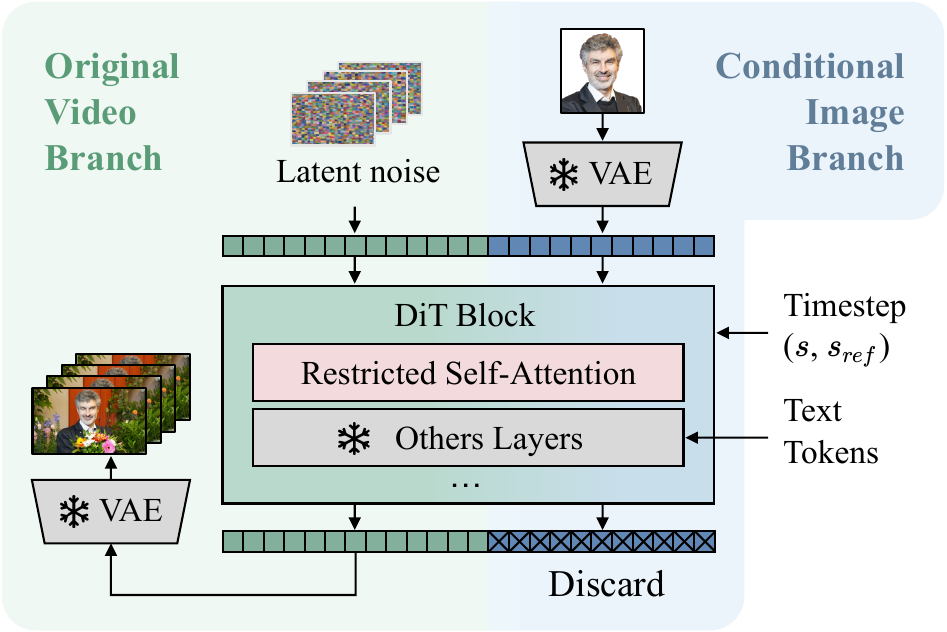}
\caption{The overview of our identity-preserving text-to-video generation framework. We introduce a conditional image branch alongside the original video branch. Given the conditional image, the VAE encoder maps it into tokens, which are concatenated with the video latent tokens and then sent to the DiT. Within the DiT blocks, identity information is incorporated into the video features through restricted self-attention.}
\label{fig:framework-overview}
\vspace{-4mm}
\end{figure}

The overall framework is illustrated in Figure~\ref{fig:framework-overview}. We use Wan2.1-14B-T2V~\cite{wan2.1} as the video generation base model, which adopts a Diffusion Transformer (DiT) architecture.
Given a reference image containing a face, we first encode it into the latent space using the pre-trained VAE encoder. The image latents undergo the same patchification and encoding procedures as the video latents. 
Then, the image tokens are concatenated with video tokens along the sequence dimension and processed jointly through successive blocks. Finally, image tokens are discarded at the final layer.

To preserve the static nature of the reference image, which serves as a conditioning input rather than undergoing denoising, we maintain its temporal invariance. This is done by fixing its timestep to zero, \ie~ $s_{ref}=0$,
where $s$ denotes the denoising timestep in diffusion.
Now that we have encoded the conditional image into the same feature space as the video, the next challenge is: How can the video features effectively refer to the image information in a way that is lightweight and easy to learn?



\vspace{-1mm}
\subsection{Restricted Self-Attention} 
\vspace{-1mm}
\label{sec:rsa}

\begin{figure*}[t]
\centering
\includegraphics[width=\textwidth]{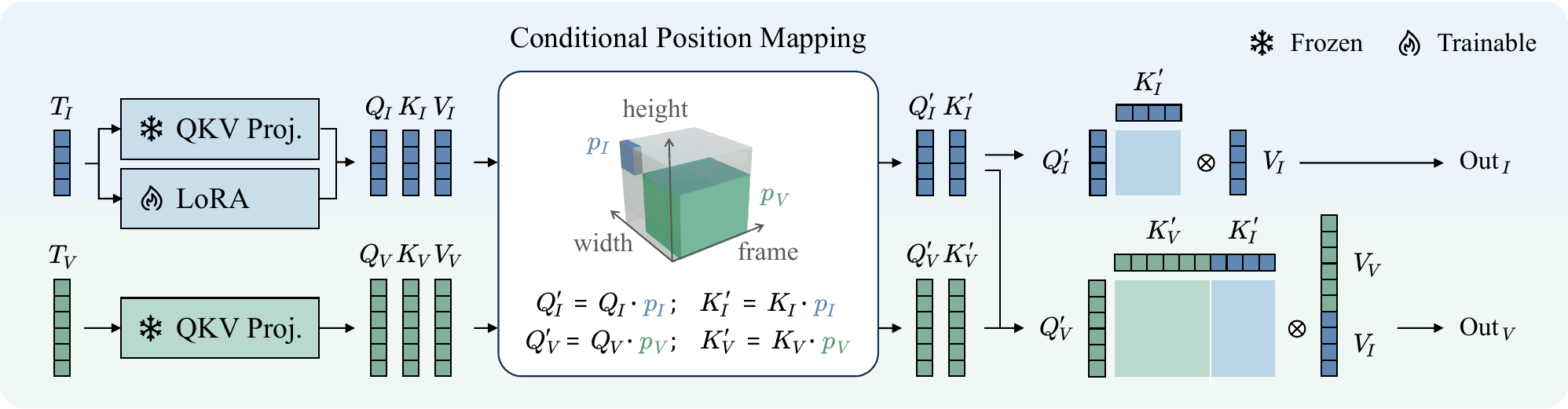}
\caption{Design of our \textbf{Restricted Self-Attention}: For the input video and image tokens, we compute their Query, Key, and Value matrices independently. Next, we apply \textbf{Conditional Position Mapping} to the Query and Key matrices. Finally, the image matrices operate independently, while the video Query performs attention using the concatenation of the image and video Key and Value matrices.}
\label{fig:framework-attention}
\vspace{-4mm}
\end{figure*}

\begin{figure}[t]
\centering
\includegraphics[width=\linewidth]{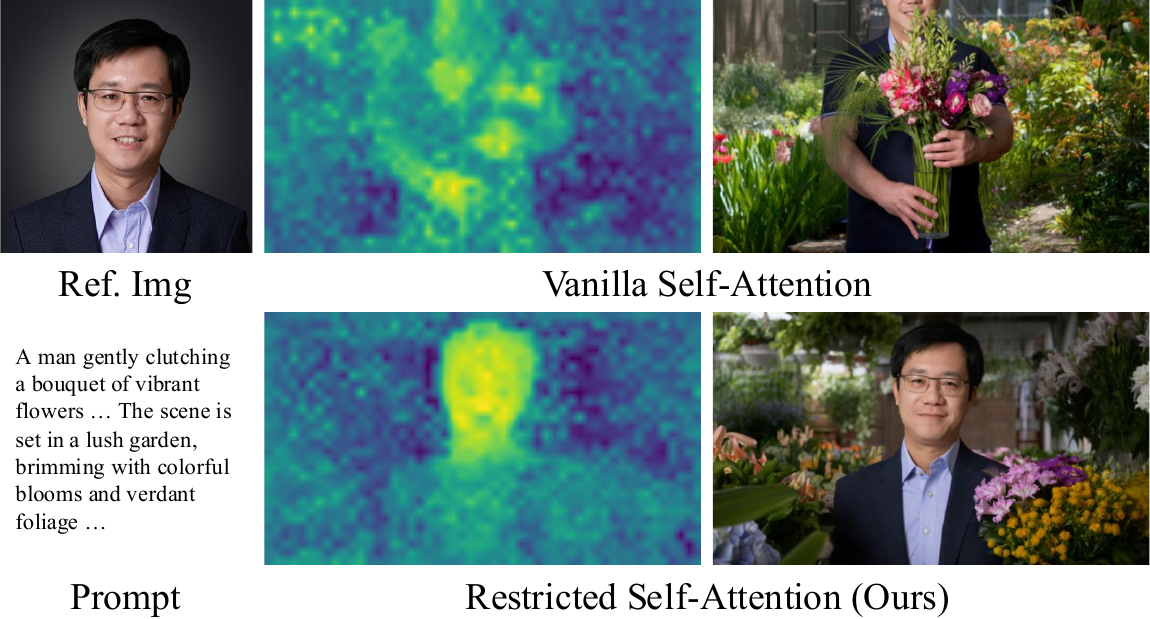}
\caption{Effect of Restricted Self-Attention (RSA). Given the reference image and the prompt (left column), we visualize attention map to reference-image tokens. Under Vanilla Self-Attention (top row), attention diffuses into background regions and the output skews toward a garden scene. With RSA (bottom row), attention concentrates on facial regions, maintaining the subject’s identity.}
\label{fig:ablation-rsa}
\vspace{-7mm}
\end{figure}

In the aforementioned DiT blocks, reference image and video tokens are processed independently through most modules (including layer normalization, cross-attention, and feed-forward networks), except for the self-attention layer. 
The self-attention layer enables information exchange among all tokens, naturally allowing video tokens to refer to identity information. 
One intuitive solution is to direct concatenate the reference-image tokens with the generated video tokens, and pass them through the Vanilla Self-Attention.
However, this approach has two main drawbacks.
First, since the reference image serves as a static condition, it should remain unaffected by the dynamic contents of the video.
Vanilla Self-Attention lets image queries attend to video contents, making it challenging to maintain the identity.
Second, this joint self-attention provides no guarantee that video tokens will actually refer to the image tokens.
The model may ignore the reference image and generate scenes without the target identity, which is shown in~\figref{fig:ablation-rsa}.


Therefore, to incorporate identity information while preserving its independence, we propose replacing the vanilla self-attention layers in DiT with a restricted version that explicitly prevents image queries from attending to video keys.
As shown in Figure~\ref{fig:framework-attention}, for a self-attention layer, we first independently compute the Query, Key, and Value for image and video tokens, denoted as $Q_I$, $K_I$, $V_I$ and $Q_V$, $K_V$, $V_V$ respectively. Then, we concatenate $K_V$ with $K_I$ and $V_V$ with $V_I$ for $Q_V$.
To enhance the model's ability to utilize identity-related information while preserving its inherent generative robustness, we incorporate Low-Rank Adaptation (LoRA) into the QKV projection of image tokens.
For analysis, we visualize attention maps specifically for video queries attending to the reference image by averaging attention maps across all DiT blocks.
As shown in~\figref{fig:ablation-rsa},
in contrast to the vanilla baseline,
our Restricted Self-Attention concentrates attention on facial regions of the reference image and yields prompt-faithful frames that preserve the subject’s identity.

Given that the timestep for the conditional image is fixed at $s_{ref}=0$, its Key and Value matrices remain constant throughout the diffusion denoising process. Therefore, during inference, we can cache $K_I$ and $V_I$ to accelerate computation, named \textbf{KV Caching}.
These matrices are computed and stored during the first denoising step, eliminating the need for redundant recalculations in subsequent steps.

\vspace{-1mm}
\subsection{Conditional Position Mapping} 
\vspace{-1mm}
\label{sec:cpm}
To effectively differentiate image and video tokens in the restricted self-attention, we use a specialized conditional position mapping strategy. Specifically, we employ 3D Rotary Positional Embedding (RoPE)~\cite{RoPE}, where all tokens associated with the reference image are assigned a distinct and dedicated coordinate space. This ensures clear separation and facilitates precise modeling of interactions between the reference image and video tokens.

For the temporal dimension, we assign a fixed temporal index of -1 to the reference image tokens, while mapping video tokens to nonnegative temporal positions. This assignment establishes image tokens as temporally invariant conditional inputs.
In this way, the model is encouraged to treat the identity information from the reference image as a constant guide throughout the entire denoising process, rather than conflating it with transient, frame-specific features in the temporal sequence.

For spatial dimensions, we employ a disjoint coordinate strategy to enforce spatial decoupling between reference image and video content. While video frames occupy coordinates within the domain $(h,w) \in [0,H_V) \times [0,W_V)$, we map reference image tokens to a dedicated coordinate subspace $[H_V, H_V+H_I) \times [W_V, W_V+W_I)$, where $H_I$ and $W_I$ represent the reference image dimensions.


Denoting $p_I$ as the coordinate for image tokens and $p_V$ for video tokens, we apply 3D RoPE as follows:
\begin{align}
    Q'_I &= Q_I \cdot p_I, & K'_I &= K_I \cdot p_I, \\
    Q'_V &= Q_V \cdot p_V, & K'_V &= K_V \cdot p_V.
\end{align}
where $\cdot$ denotes the Hadamard product. The restricted self-attention outputs are computed as:
\begin{align}
    \text{Out}_I &= \text{Attention}(Q'_I, K'_I, V_I), \\
    \text{Out}_V &= \text{Attention}(Q'_V, [K'_V, K'_I], [V_V, V_I]),
\end{align}
where $[\cdot, \cdot]$ denotes concatenation.

\begin{figure}[t]
\centering
\includegraphics[width=\linewidth]{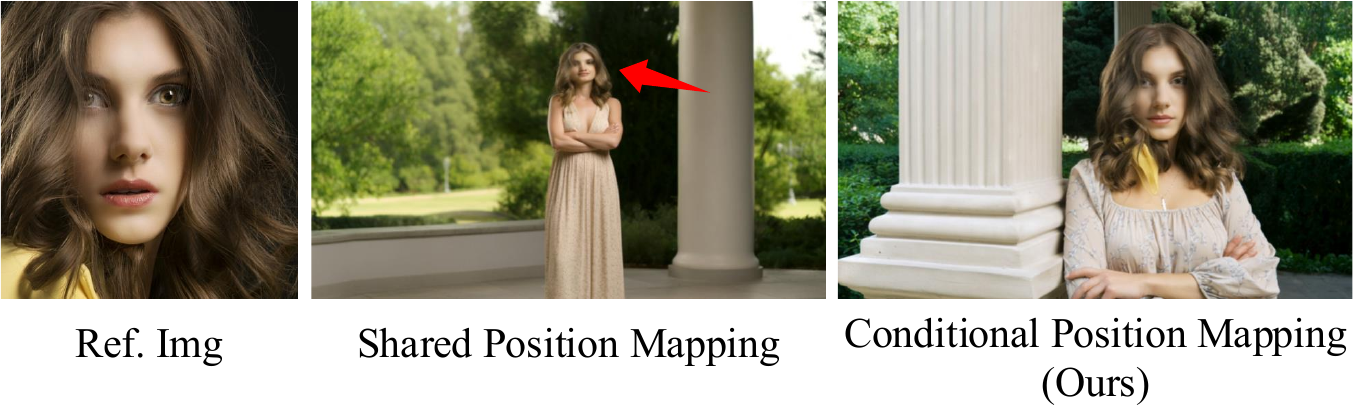}
\caption{Effect of Conditional Position Mapping (CPM). Compared with Shared Postion Mapping, our CPM, where the reference tokens are mapped to a disjoint spatial space, better preserves the pretrained positional prior and yields more stable scenes.}
\label{fig:ablation-pos}
\vspace{-5mm}
\end{figure}

This non-overlapping spatial allocation achieves two main goals through geometric separation. 
By separating the reference tokens from the video coordinate grid, the design naturally reduces false spatial correlations and better preserves the backbone’s pretrained positional prior.
Compared with Shared Position Mapping, our Conditional Position Mapping generates more reliable video with more stable proportions, which is shown in~\figref{fig:ablation-pos}.
At the same time, this separate coordinate system maintains the reference image’s semantic meaning by keeping it as a global identity prior. Consequently, the model is guided to focus on extracting overall semantic features from the reference tokens, rather than treating them as spatially localized patterns that need to align positionally with the video content.

\vspace{-1mm}
\subsection{Dataset Collection and Processing}
\label{sec:dataset}
\vspace{-1mm}

We construct a human-centric video dataset containing 2,000 high-resolution sequences from publicly available sources.
The dataset guarantees a diverse and comprehensive representation, comprising various ethnic groups, age ranges, gender identities, and a wide array of actions.
Using the VILA~\cite{lin2024vila} multimodal captioning framework, we automatically generate dense textual annotations for each video, establishing strong text-video alignment.

To align the dataset with the pre-training distribution of our video generation base model~\cite{wan2.1} and to mitigate potential degradation in generation quality, we preprocess the videos as follows: each video is resampled to 25 FPS, then cropped and resized to a resolution of 832$\times$480 pixels. From these processed videos, we randomly sample clips of 81 consecutive frames for training.

For each video clip, the corresponding reference facial image is extracted from the original, pre-resampled video. The extraction pipeline is as follows:
\begin{enumerate}
\item 5 frames are randomly selected from the original video.
\item Faces are detected and cropped using RetinaFace\cite{Deng2020CVPR}.
\item The cropped face images are resized to 512$\times$512 pixels.
\item BiSeNet~\cite{BiseNet} is used for face parsing, and the background is replaced with a solid white color to prevent any leakage of background information.
\end{enumerate}

\begin{figure}[t]
    \centering
    \includegraphics[width=\linewidth]{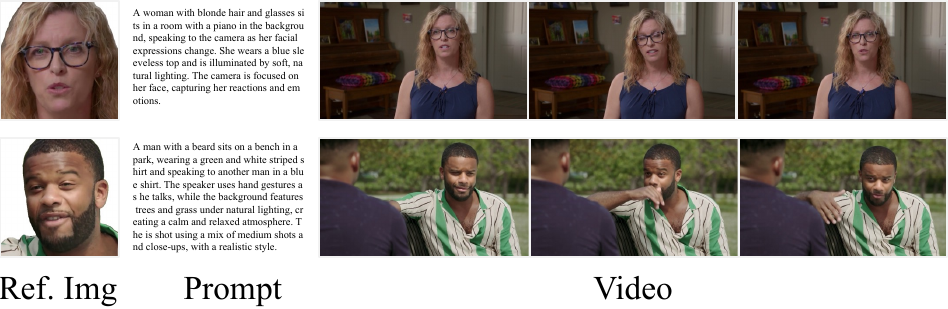}
    \caption{Examples from our human-centric video dataset.}
    \label{fig:dataset}
\vspace{-3mm}
\end{figure}

\noindent
Examples of the final image-text-video pairs for our training can be found in Figure~\ref{fig:dataset}.

\begin{table}[!t]
\centering
\footnotesize
\renewcommand{\arraystretch}{1.1}
\setlength{\tabcolsep}{4.0pt}
\caption{
    \textbf{Quantitative comparison with state-of-the-art identity-preserving video generation methods.}
    We evaluate across three key metrics: Face Similarity, Naturalness, and Prompt Following.
    For all metrics, higher values indicate better performance.
    The best and second-best results in each column are highlighted in \textbf{bold} and \underline{underlined}, respectively.
}
\vspace{-2mm}
\begin{tabular}{l|cccc}
\toprule
\textbf{Method} & \begin{tabular}[c]{@{}c@{}}\textbf{Face}-\\ \textbf{Similarity}\end{tabular} $\uparrow$ & \textbf{Naturalness} $\uparrow$ & \begin{tabular}[c]{@{}c@{}}\textbf{Prompt}-\\ \textbf{Following}\end{tabular}$\uparrow$ \\
\midrule
\multicolumn{4}{l}{\textit{Closed-Source Methods}}    \\
Kling &  0.410 & \underline{3.900} & 19.921 \\
Hailuo & 0.577 & 3.750 & \textbf{20.649} \\
Pika-2.1 & 0.323 & 3.644 & \textbf{20.649} \\
Vidu-2.0 & 0.361 & 3.600 & 18.998 \\
\midrule
\multicolumn{4}{l}{\textit{Open-Source Methods}} \\
ID-Animator \cite{ID-Animator}& 0.316 & 3.211 & 16.677 \\
SkyReels-A2-P14B \cite{SkyReels} & 0.546 & 3.411 & 19.110 \\
EchoVideo \cite{EchoVideo}&0.487 & 3.456 & 19.263 \\
ConcatID-CogVideoX \cite{Concat-ID}&0.439 & 3.372 & 19.359 \\
ConcatID-WAN \cite{Concat-ID}& 0.501 & 3.650 & 19.671 \\
Hunyuan-Custom \cite{HunyuanCustom}& 0.622 & 3.367 & 19.853 \\
ConsistID \cite{consistid}& 0.432 & 3.233 & 20.552 \\
VACE-P1.3B \cite{vace}& 0.180 & 3.567 & 20.591 \\
VACE-1.3B \cite{vace}& 0.223 & 3.611 & 20.527 \\
VACE-14B \cite{vace}& \underline{0.647} & 3.728 & 19.520 \\
Phantom-1.3B \cite{liu2025phantom} & 0.440 & 3.567 & 20.364 \\
Phantom-14B \cite{liu2025phantom} &  0.519 & 3.828 & 20.476 \\
\textbf{Stand-In (Ours)} &\textbf{0.724} & \textbf{3.922} & \underline{20.594} \\
\bottomrule
\end{tabular}
\label{tab:quantitative_comparison_simplified}
\vspace{-4mm}
\end{table}

\begin{figure*}[t]
\centering
\includegraphics[width=\textwidth]{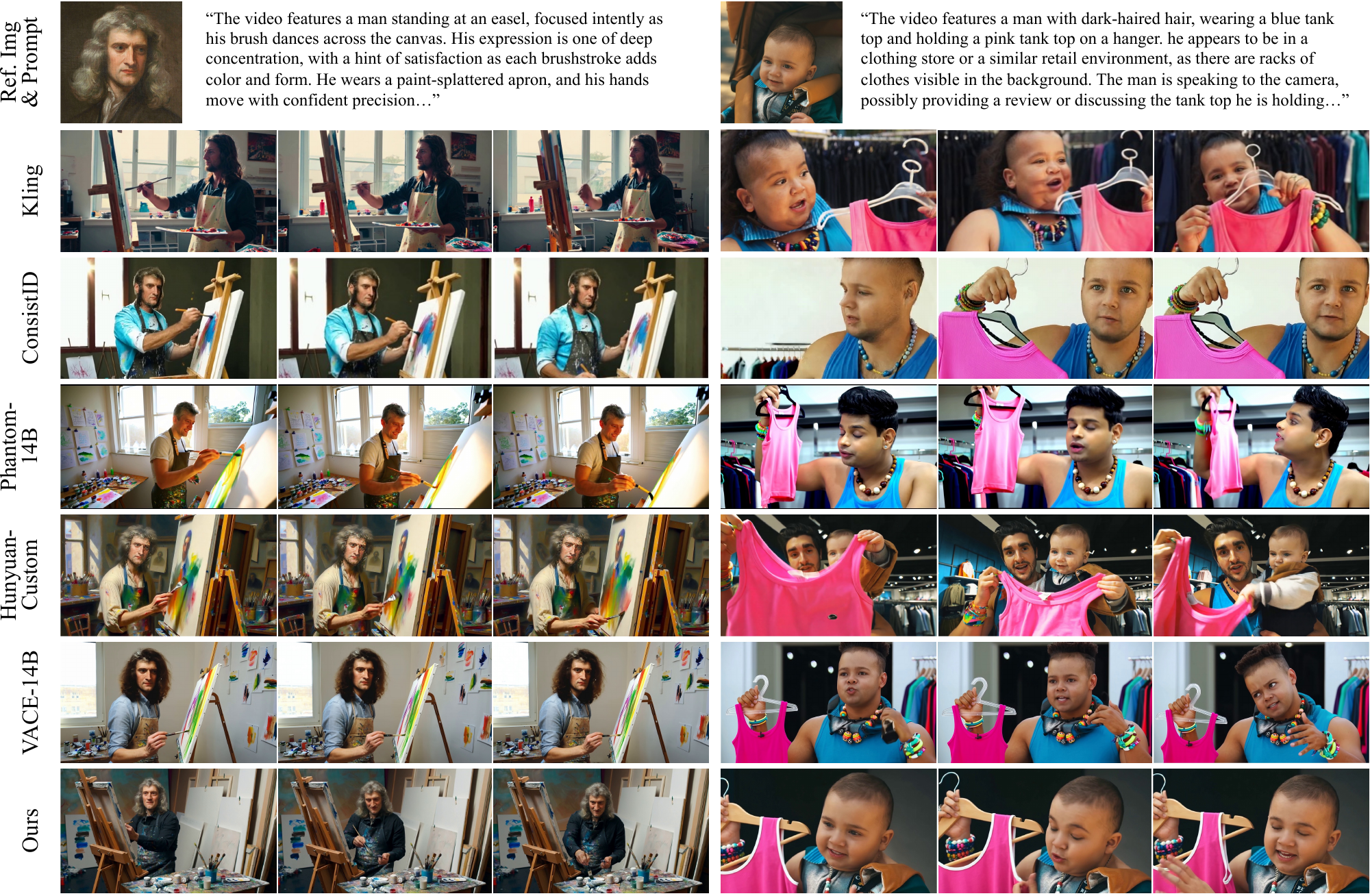}
\caption{Comparison on identity-preserving video generation. Please refer to the supplementary material for full prompts.}
\label{fig:comparison-main}
\vspace{-2mm}
\end{figure*}

\vspace{-2mm}
\section{Experiments}
\vspace{-1mm}
\subsection{Implementation Details} 





We adopt LoRA with rank 128, applied only to the QKV projections for image tokens in each DiT block. For the 14B-parameter Wan2.1 model, this adds just 153M trainable parameters (1\% of the base model), increasing feed-forward time by 3.6\% and FLOPs by 2.6\%. During inference with KV caching, overhead is minimal: runtime rises by only 2.3\% and FLOPs by 0.07\% compared to the video generation base model. This negligible cost shows our identity-preserving method is \textbf{lightweight}.

\begin{table}[!t]
\centering
\small
\setlength{\tabcolsep}{5pt} 
\caption{
    \textbf{User study results for subjective evaluation.} The best and second-best results in each column are highlighted in \textbf{bold} and \underline{underlined}, respectively.
}
\begin{tabular}{lcc}
\toprule
\textbf{Method} & \textbf{Face Similarity} $\uparrow$ & \textbf{Video Quality} $\uparrow$ \\
\midrule
Hunyuan-Custom\cite{HunyuanCustom} & \underline{3.34} & 2.92 \\
VACE-14B \cite{vace}& 3.00 & 3.07 \\
Phantom-14B \cite{liu2025phantom}& 2.37 & 2.92 \\
ConsistID \cite{consistid}& 2.25 & 2.46 \\
Kling & 2.21 & \underline{3.09} \\
\textbf{Stand-In (Ours)} & \textbf{4.10} & \textbf{4.08} \\
\bottomrule
\end{tabular}
\label{tab:user_study}
\end{table}
The model is trained over 3000 steps on Nvidia H20 GPUs with a batch size of 48.
For inference, BiSeNet is adopted as an automatic preprocessing step.
\begin{figure}[t]
\centering
\includegraphics[width=\linewidth]{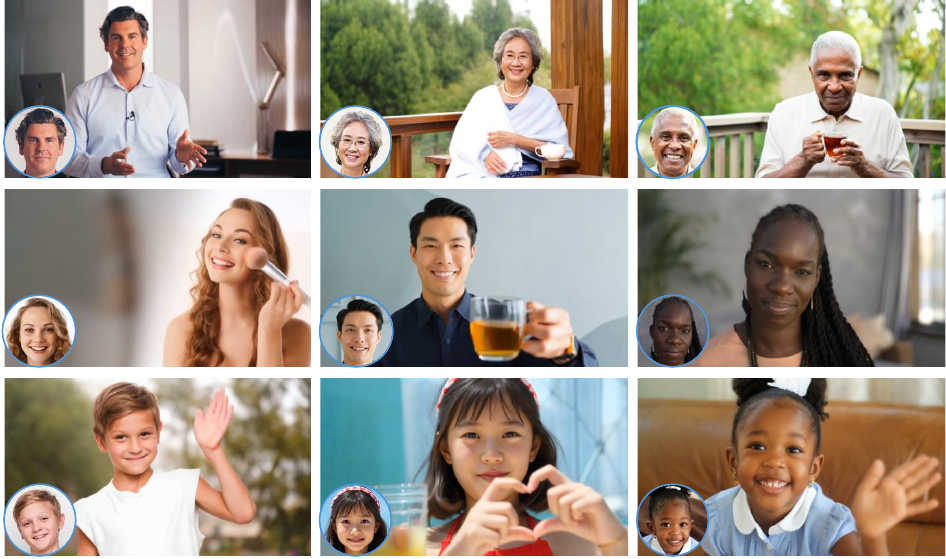}
\caption{Our model generalizes to unseen ordinary individuals across diverse ethnicities and age groups, despite being trained with only $\sim$1\% additional parameters and just 2000 training pairs.}
\label{fig:generalize}
\vspace{-6mm}
\end{figure}

\begin{figure*}[!t]
\centering
\includegraphics[width=\linewidth]{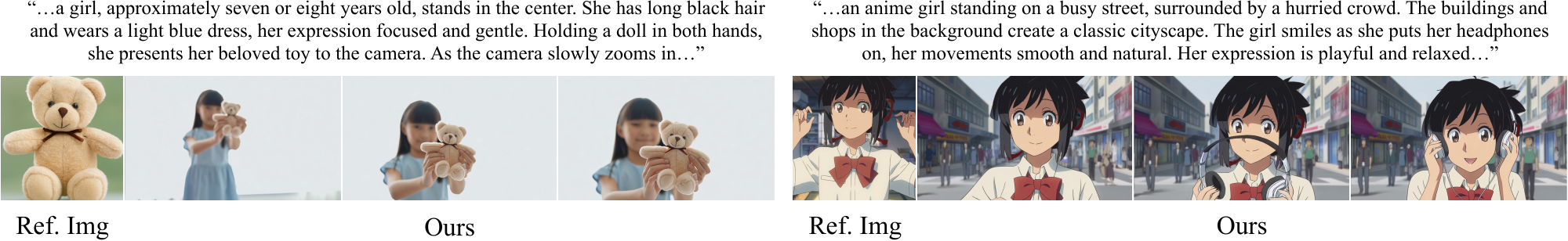}
\caption{Our results on subjects other than real-person. Please refer to the supplementary material for full prompts.}
\label{fig:subject}
\end{figure*}

\begin{figure*}[t]
\centering
\includegraphics[width=\linewidth]{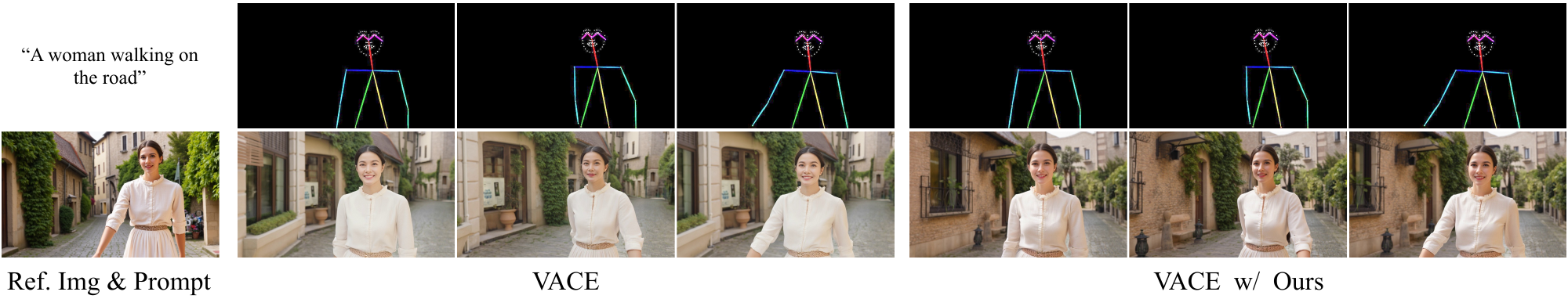}
\caption{Comparison on pose-guided video generation against VACE.}
\label{fig:pose-guided_comp_vace}
\end{figure*}

\begin{figure*}[t]
\centering
\includegraphics[width=\linewidth]{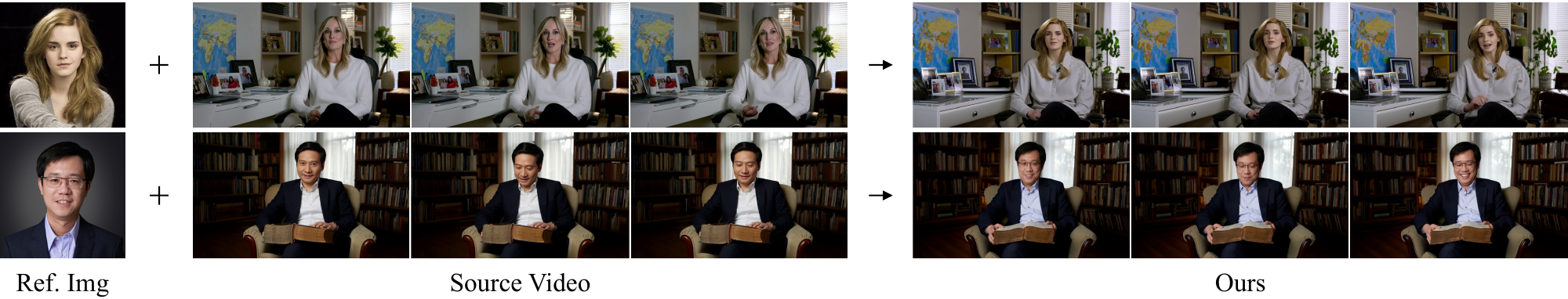}
\caption{Application of our model in video face swapping.}
\label{fig:swap}
\end{figure*}

\begin{figure}[t]
\centering
\includegraphics[width=\linewidth]{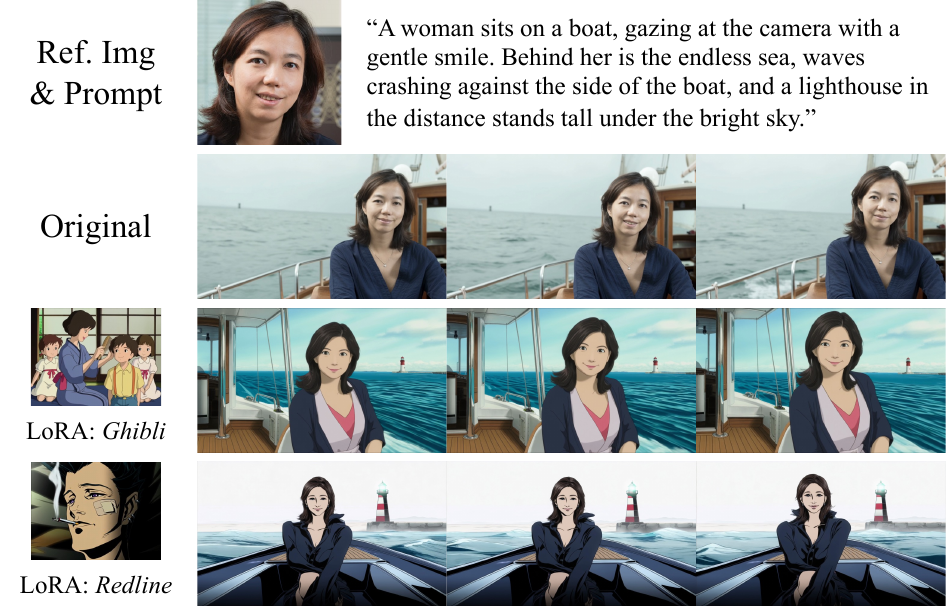}
\caption{Our model applied with stylization LoRA.}
\label{fig:lora}
\end{figure}
\vspace{-1mm}
\subsection{Quantitative Analysis}
\vspace{-1mm}

To evaluate identity preservation and visual quality, we use the two most important and heavily weighted evaluation metrics from the OpenS2V benchmark~\cite{yuan2025opens2vnexusdetailedbenchmarkmillionscale}: facial similarity and naturalness. To evaluate the relevance between generated video and textual description, we use X-CLIP~\cite{xclip}, a pre-trained video-text multimodal model. Results are shown in Table~\ref{tab:quantitative_comparison_simplified} and Figure~\ref{fig:comparison-main}.
As shown in~\figref{fig:generalize}, despite being trained with only $\sim$1\% additional parameters and just 2000 training pairs, our model also demonstrates strong generalization across individuals of diverse ethnicities and age groups.

\paragraph{Face Similarity}
This metric evaluates the model's ability to maintain identity consistency. It is calculated as the average cosine similarity between the CurricularFace embeddings of the reference image and the faces detected in the generated video frames. As shown in Table~\ref{tab:quantitative_comparison_simplified}, our proposed method, Stand-In, achieves a score of 0.724, outperforming all other compared methods. This result demonstrates the effectiveness of our approach in generating facial features that remain highly consistent with the source identity.
\vspace{-4mm}
\paragraph{Naturalness}
This metric is primarily designed to evaluate the naturalness of the generated videos. It leverages GPT-4o to approximate human judgment of video realism, taking into account factors such as physical plausibility and the absence of noticeable AI artifacts. Following the OpenS2V protocol, a holistic score ranging from 1 to 5 is assigned. In this evaluation, our method achieves a score of 3.922, demonstrating that the improvements in identity fidelity are achieved without compromising the overall visual realism of the generated videos.
\vspace{-4mm}
\paragraph{Prompt Following}
As shown in Table~\ref{tab:quantitative_comparison_simplified}, our method ranks second among all compared methods and first among open-source methods. This result demonstrates that our identity-preserving ability can be achieved without hurting the prompt-following performance.
\vspace{-1mm}
\subsection{User Study}
\vspace{-1mm}

The user study involves 20 participants. We randomly select 10 test videos from the benchmark and ask the participants to rate each video across two dimensions: \textit{facial similarity} and \textit{video quality}. The latter dimension encompasses aspects such as naturalness, visual aesthetics, and alignment with the provided text descriptions. Ratings are given on a 5-point scale (1 to 5), and the final scores for each dimension are obtained by averaging the ratings across all participants and test videos. As shown in Table~\ref{tab:user_study}, our method outperforms the comparative methods.

\vspace{-1mm}
\subsection{Plug-and-Play to Other Applications}
\vspace{-1mm}
\paragraph{Subject-Driven Video Generation}

Although trained only with human data, our framework can be zero-shot applied to non-human subjects without any additional fine-tuning. This is because we use the pretrained VAE and video generation model to extract rich features, and learn alignment through paired data and an effective attention mechanism. This zero-shot ability can hardly be achieved by traditional identity-preserving methods relying on face encoders.
As shown in Figure~\ref{fig:subject}, our method exhibits strong subject consistency on the teddy bear object and the cartoon character.
\vspace{-3mm}
\paragraph{Pose-Guided Video Generation}

The proposed conditional image branch is designed based on the LoRA module, which ensures inherent compatibility with other DiT-based models. To validate this, we conducted experiments on the pose-guided video generation task using the VACE framework \cite{vace}. As illustrated in Figure \ref{fig:pose-guided_comp_vace}, integrating our method significantly improves the facial identity similarity in the generated videos. This not only demonstrates the plug-and-play nature of our approach but also highlights its robustness in preserving identity consistency.
\vspace{-3mm}
\paragraph{Video Stylization}
By applying our framework in conjunction with video stylization LoRA, we demonstrate its ability to achieve effective style transfer while maintaining strong identity consistency. As shown in Figure~\ref{fig:lora}, our method successfully renders the artistic style as well as preserving the facial features of the reference image, further showing its versatility and robustness.

\paragraph{Video Face Swapping}


Our framework is also capable of video face swapping, which can be achieved via zero-shot inpainting~\cite{inpainting}. Figure~\ref{fig:swap} shows that our method not only achieves high-quality facial identity transfer but also maintains strong temporal consistency across frames, resulting in coherent and high-quality videos.

\subsection{Ablation Study}

\paragraph{Effectiveness of Restricted Self-Attention (RSA)}

To verify the role of RSA, we replaced it with Vanilla Self-Attention (VSA), which concatenates the reference-image and video tokens and allows full bidirectional information flow.
Because VSA treats the reference tokens as part of the dynamic video context, the model often fails to refer to the reference image.
As shown in~\figref{fig:ablation-rsa}, this naive design leads to diffused attention and unreliable identity preservation.
In contrast, RSA constrains attention so that image queries cannot attend to video tokens. This design keeps the reference representation static while still providing identity cues.
As shown in~\tabref{tab:ablation_study}, RSA improves face similarity from 0.422 to 0.724 and also slightly enhances naturalness, confirming its essential role in maintaining both identity consistency and visual quality.

\begin{table}[t]
\centering
\small
\setlength{\tabcolsep}{5pt} 
\renewcommand{\theadfont}{\bfseries} 
\caption{
    \textbf{Ablation study on the core components of our method.} Replacing Restricted Self-Attention (RSA) with Vanilla Self-Attention (VSA) or Conditional Position Mapping (CPM) with Shared Position Mapping (SPM) degrades both Face Similarity and Naturalness. 
}
\begin{tabular}{l c c c}
\toprule
\thead{Method} & \thead{Face Similarity $\uparrow$~~} & \thead{Naturalness $\uparrow$}  \\
\midrule
RSA $\rightarrow$ VSA & 0.422 & 3.855 \\
CPM $\rightarrow$ SPM & 0.536 & 3.755 \\
\midrule 
\textbf{Full Model} & \textbf{0.724} & \textbf{3.922} \\
\bottomrule
\end{tabular}
\label{tab:ablation_study}
\vspace{-4mm}
\end{table}

\paragraph{Effectiveness of Conditional Position Mapping (CPM)}



We further investigate the effect of CPM by Replacing CPM with a Shared Position Mapping (SPM), which places reference and video tokens within the same positional coordinate system.
As illustrated in~\figref{fig:ablation-pos}, SPM breaks the pretrained positional prior of the backbone model, resulting in unstable spatial layouts.
By geometrically separating the coordinate spaces, CPM preserves the pretrained positional prior while maintaining the reference as a global identity prior.
In~\tabref{tab:ablation_study}, CPM increases face similarity from 0.536 to 0.724 and improves naturalness from 3.755 to 3.922, demonstrating its effectiveness in stabilizing video synthesis and enhancing identity fidelity.

\section{Conclusion}
We propose \textbf{Stand-In}, a lightweight, plug-and-play framework for high-fidelity, identity-preserving video generation.
We introduce a conditional image branch into a pre-trained video generation model, and propose a restricted attention mechanism with conditional positional encoding to enable cross-branch information exchange. Despite training only 1\% of the model's additional parameters on a limited dataset of 2,000 pairs, our approach achieves high-quality video generation while maintaining strong identity fidelity.
Experimental results demonstrate that Stand-In achieves state-of-the-art performance in identity-preserving text-to-video generation. Furthermore, it exhibits excellent performance on other tasks, including pose-guided video generation, stylization, and face swapping, proving its strong compatibility and broad application potential.



\section{Acknowledgment}
This work was supported in part by the National Natural Science Foundation of China (62306153, 62225604), Tianjin Natural Science Foundation Project (25ZXRGGX00290, 24JCJQJC00020, 25JCQNJC01390),
the Young Elite Scientists Sponsorship Program by CAST (YESS20240686), the Fundamental Research Funds for the Central Universities (Nankai University, 63253223, 63253219), and Shenzhen Science and Technology Program (JCYJ20240813114237048). The computational devices is supported by the Supercomputing Center of Nankai University (NKSC).
{
    \small
    \bibliographystyle{ieeenat_fullname}
    \bibliography{main}
}


\end{document}